# 3D SEQUENTIAL IMAGE MOSAICING FOR UNDERWATER NAVIGATION AND MAPPING

E. Nocerino [1]*, F. Menna [2], B. Chemisky [1], P. Drap [1]

[1] LIS UMR 7020, Aix-Marseille Université, CNRS, ENSAM, Université De Toulon, Marseille, 13397, France – (erica.nocerino, pierre.drap)@univ-amu.fr, bertrand.chemisky@etu.univ-amu.fr
[2] 3D Optical Metrology (3DOM) unit, Bruno Kessler Foundation (FBK), Trento, Italy - fmenna@fbk.eu

**KEY WORDS:** Underwater navigation, image stitching, image mosaicing, SLAM, visual odometry

**ABSTRACT:**

Although fully autonomous mapping methods are becoming more and more common and reliable, still the human operator is regularly employed in many 3D surveying missions. In a number of underwater applications, divers or pilots of remotely operated vehicles (ROVs) are still considered irreplaceable, and tools for real-time visualization of the mapped scene are essential to support and maximize the navigation and surveying efforts. For underwater exploration, image mosaicing has proved to be a valid and effective approach to visualize large mapped areas, often employed in conjunction with autonomous underwater vehicles (AUVs) and ROVs. In this work, we propose the use of a modified image mosaicing algorithm that coupled with image-based real-time navigation and mapping algorithms provides two visual navigation aids. The first is a classic image mosaic, where the recorded and processed images are incrementally added, named 2D sequential image mosaicing (2DSIM). The second one geometrically transform the images so that they are projected as planar point clouds in the 3D space providing an incremental point cloud mosaicing, named 3D sequential image plane projection (3DSIP). In the paper, the implemented procedure is detailed, and experiments in different underwater scenarios presented and discussed. Technical considerations about computational efforts, frame rate capabilities and scalability to different and more compact architectures (i.e. embedded systems) is also provided.

## 1. INTRODUCTION

Image stitching is the procedure of registering together multiple images to obtain a unified 2D representation of a physical space, whose most common result is a panorama, i.e. a wide-angle view of the scene. The process is also known as image mosaicing or mosaicking, word deriving from the Latin 'mosaicum', a planar composition of coloured tiles arranged together in a pattern. The main difference between a 'classic' mosaic and photo mosaic is that the latter automatically produces a composite image, where the single images have a certain overlap. As well-known, the overlap area is used to estimate the geometric transformation to register together the adjacent images.

Image mosaicing finds many applications in different fields, as summarised by Adel et al. (2014) and Ghosh and Kaabouch (2016): high-resolution photomosaics or panorama creation, augmented reality, resolution enhancement or super-resolution, motion detection and tracking, image stabilization, video indexing and compression, to cite a few.

As demonstrated by the huge amount of research work on the topic, image mosaicing is an established approach for autonomous or remotely mapping and real-time visualization. The technique has been indeed utilized as aid for robot path planning, navigation, and mapping on land (Kelly, 2000; Lucas et al., 2010; Wang et al., 2019) and underwater (Eustice, 2005; Gracias et al., 2003); for environmental monitoring through geo-referenced video registration without a digital elevation model – DEM from unmanned aerial vehicles (Zhu et al., 2005); with video acquired with large format aerial vehicles (Molina and Zhu, 2014), for surveillance (Yang et al., 2015) and tracking of moving objects (Linger and Goshtasby, 2014); for constructing an overview of a target area with different sensors (RGB and/or thermal cameras) with and without metadata from the GPS and inertial navigation system (INS) from small-scale UAV (Yahyanejad, 2013); for supporting image interpretation and navigation in medical applications with microscopes (Loewke et al., 2010).

In this work we describe a novel optimized approach of image mosaicing primarily tailored for the observation and real time mapping of the seabed. Integrated in a visual odometry (VO), SLAM or incremental SfM (iSfM) process, the proposed image mosaicing is mainly thought to be provided to the pilot of a remotely operated vehicle (ROV) or a diver to check that the area of interest is covered, and the mapping is properly performed. The image planes are incrementally mosaiced and visualised also in 3D along with the conventional output provided by VO/SLAM or iSfM, i.e. camera pose and sparse point cloud. Thanks to its modularity, as detailed in the followings, the implemented method could be further adapted and integrated into navigation and planning modules for autonomous navigation vehicles (AUVs).

The manuscript is structured as follows. First, an overview of image mosaicing is provided, with special emphasis on studies focused on underwater applications. Then, the implemented procedure is detailed and considerations about computational efforts, frame rate capabilities and scalability to different and more compact architectures (i.e. embedded systems) are provided. The paper concludes with experiments in different scenarios and outlines some future research avenue.

---

\* Corresponding author

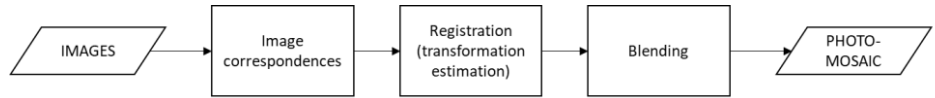

Figure 1. The standard workflow in image mosaicing (after Adel et al., 2014 and Ghosh and Kaabouch, 2016)

## 2. OVERVIEW ON IMAGE MOSAICING AND ITS APPLICATION FOR UNDERWATER MAPPING

Image mosaicing entails sequential steps, as summarised in Figure 1. The image correspondence problem can be solved in a number of different ways, classifiable primarily into frequency and spatial domain-based, with the latter including the two well-known area and feature-based methods (Ghosh and Kaabouch, 2016).

Once correspondences are established, the information derived is used to estimate the mathematical transformation between the images, aligning them into a common reference frame (for example, the first image). Several transformations are possible, corresponding to different parametric motion models i.e. relation between two images, which ranges from 2D transformations to planar perspective models or mapping to non-planar cylindrical surfaces (Szeliski, 2010). The commonly adopted mathematical model in image stitching is an eight-parameter homography, i.e. the 2D projective transformation which describes how a planar surface is projected onto the image plane. The homography assumption optimally holds when the scene is nearly planar or the motion between the images is purely rotational, meaning that the camera has been rotated around its optical centre without any translational displacement. Under specific circumstances, as when, for examples images are taken with a levelled camera (or the tilt angle is known), the images can be first mapped or wrapped using different types of projections (cylindrical, stereographic, equirectangular, etc.) (Cambridge in colour, 2020) and then aligned using a specific motion model (e.g., a rotation of the camera is a translation in the cylindrical space) (Chuang, 2007). After the transformation is estimated, the images are transformed projected into the common space, and, then, stitched together in a bigger image. When many images are stitched together through a sequential process, residual errors can accumulate (drift) and a global optimization is performed, which provides updated transformation parameters. The global registration can be estimated through different methods, such as bundle adjustment (Szeliski, 2010). After this step, the overlapping areas between the images are merged through blending algorithms, which aim at reducing any photometric residual misalignments and providing a seamless final photo-mosaic.

Image mosaicing has been widely investigated for underwater exploration (Singh et al., 2004; Prados et al., 2014), proving to be an effective and fast approach to derive a representation of wide areas in an adverse environment (Elibol et al., 2011). Coupled with the use of autonomous underwater vehicles (AUV) or remotely operated vehicles (ROV), underwater image mosaics have been exploited for navigation, localization of areas of interest and detection of temporal changes (Elibol et al., 2011). Challenges are posed by the unfavourable environmental conditions of the underwater setting, i.e. scattering due to particles suspension, movements of marine fauna and flora, light absorption, refraction, as well as by the difficulties of remote operation and therefore the instability of the image acquisition platform, i.e. variability of the distance to the scene and its relative speed. These constraints may seriously affect the image registration step, to the point of failure or accumulation of not negligible errors.

### 2.1 Computational and visualization constraints for real-time underwater navigation and mapping

While real time navigation and mapping algorithms have become a powerful tool to estimate the ego-motion of the underwater vehicle in three dimensions, the sparsity of the reconstructed point cloud (mainly for computational reasons) does not allow 1) to understand the effective areas photographed and 2) limits online analyses and understanding of the observed scenery underwater.

In underwater real time applications, image acquisition and transfer frame rates have to guarantee efficient and fast processing times for autonomous navigation (AUVs) or a smooth visualization for remotely guided navigation (ROVs). Indeed, the delay between the visual information transmitted to the pilot, the control action he undertakes and the actual result in terms of the response of the vehicle must be minimised as much as possible. The frame rate is also function of the operational distance, the camera field of view and vehicle speed. From the authors' experience, in offshore industrial inspection tasks for example, for a vehicle to object distance less than 2 m and an operational speed lower than 0.5 m/s, a frame rate of 10 Hertz for the transmitted video, 5 Hertz for trajectory and 0.5 Hertz for 3D scene rendering updates are processing frequencies well tolerated by the pilot of an ROV.

Moving from these considerations, the approach herein described and summarised in Figure 2 has been developed.

## 3. THE DEVELOPED PROCEDURE

The method assumes that the camera is pre-calibrated. An essential step in the proposed approach is the VO/SLAM/iSfM step, where the image correspondence extraction and matching is performed, image pose is computed and a sparse point cloud is iteratively generated. This is crucial, as the sparse point cloud is used to fit a local reference plane. The estimated image poses and reference planes are used to compute the projective transformation and rectify the images. Two products are generated: the first is an image mosaic, where the images are incrementally added (2D sequential image mosaicing - 2DSIM). As additional output, the image planes are projected as point clouds in the 3D space providing an incremental point cloud mosaicing (3D sequential image plane projection - 3DSIP).

Scaling is guaranteed either through the use of stereo-camera or the use of a laser pointer. Attitude and heading reference system (AHRS) can be also integrated to provide the vertical information.

Each step of the method is built in four blocks or modules: (1) the VO/SLAM/iSfM block; (2) plane fitting; (3) sequential image stitching and image plane projection in object space; (4) the visualization block.

### 3.1 Overview of the system architecture

The implemented algorithmic blocks are intended as part of a larger system architecture for underwater photogrammetry purposes, employing different platforms, such as AUVs, ROVs and divers.

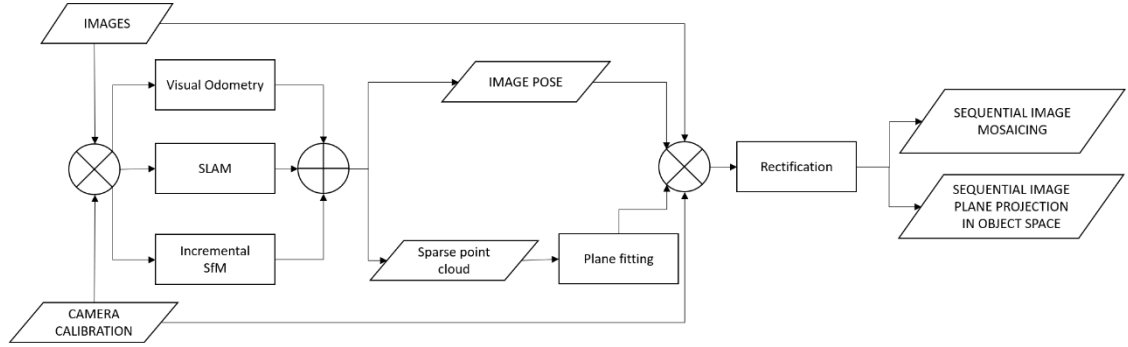

Figure 2. Developed image stitching workflow

| NAME | | COMEX POOL | NIWA T2-2017 | Moorea RESILIENCE SITE plots 4-7 | XLENDI |
|---|---|---|---|---|---|
| ACQUISITION PLATFORM | | Diver | Diver | Diver | SUBMARINE |
| CAMERA | | ORUS 3D underwater photogrammetry system | Sony A7sII | Panasonic DC-GH5S | ROV-3D underwater photogrammetry system |
| ACQUISITION MODE | | Still images | Video | Still images | Still images |
| IMAGE RESOLUTION (pixel) | Original | 1936 x 1456 | 1920 x 1080 | 3680 x 2760 | 6576 x 4384 |
| | Processed | 968 x 728 (1/4 x full resolution) | 960 x 540 (1/4 x full resolution) | 920 x 690 (1/16 x full) resolution | 1644 x 1096 (1/16 x full) resolution |
| FRAME RATE (fps) | Original | 10 | 25 | 0.5 | 2 |
| | Processed | 0.5 | 2 | 0.5 | 2 |
| PROCESSED IMAGES | | 121 | 300 | 580 | 80 |
| AREA COVERED (m) | | 1.2 x 30 | 4 x 21 | 15 x 30 | 1.5 x 12 |
| DEPTH (m) | | 3 | 21 | 12 | 110 |
| SEAFLOOR CHARACTERISTICS | | Flat | Flat with irregularities and living organisms | Hilly and rough | Flat with irregularities (archaeological assets) |
| ACQUISITION STRATEGY | | Transect / single strip | Transect / single strip | Plot / multiple round trips | Transect / single strip |
| APPLICATION DOMAIN | | Subsea metrology | Marine ecology | Marine ecology | Archaeology |

Table 1. Datasets characteristics

In this architecture a single or multi-camera array is connected to a microcomputer in charge of the navigation solution based on visual methods in real time (possibly integrating other sensors such as attitude and heading reference systems – AHRS, doppler velocity log – DVL, sonar altimeter). This system component is then connected to a control and visualization unit that may consist of a waterproof tablet PC attached to the sensor arrays underwater or, when using an ROV for subsea metrology inspections, be a more powerful surface computer on the support vessel.

To adapt to a wide range of vehicles and applications, we propose a modular architecture as described in the ROV3D project (Drap et al., 2015b). Indeed, remote operations in the underwater environment require data links between the surface and the vehicle, which are often critical in terms of throughput and robustness. The proposed architecture provides data storage and part of the processing onboard the underwater unit, so that the data volume to be transferred to a visualisation unit, typically on the surface, is significantly reduced, making it possible to maintain operability even in the case of low bandwidth of the data link e.g 50Mb to 100Mbps rate.

This architecture has the advantage of easily adapting to a SCUBA diving application where the visualization and mosaic module can be embedded on a submersible tablet. If the solution is operated from an ROV, the data to be transmitted through the umbilical will be minimized and only one image every 2 seconds will have to be sent to the surface to the visualization and mosaicing module. Furthermore, this architecture allows an implementation on an AUV by directly exploiting the results of the VO/SLAM/iSfM processing for integration into the vehicle's navigation and planning module.

### 3.2 The VO/SLAM/iSfM module

This block takes as input the image stream and provides as output the image poses and the sparse point could of tie points.

The current implementation includes a VO with windowed bundle adjustment (wBA). The process starts when a first set on n (three or five in the experiments) images are acquired, then a two-step procedure is followed: (i) for each new image, a relative orientation is computed; (ii) a sliding window selects the last n images and a block adjustment is performed (Figure 3a). The previous blocks are kept fixed, to preserve a common reference system and scale. If the camera motion is too abrupt and the sequential image orientation fails, an alert is provided to the user, the process stops, and the image acquisition should be started again. Loop-closure is currently not implemented.

The VO-wBA is based on the method presented in Menna et al. (2019a, 2019b) and here simulated through a prototypal approach implemented within the Agisoft Metashape Python API (Metashape Python Reference, 2020).

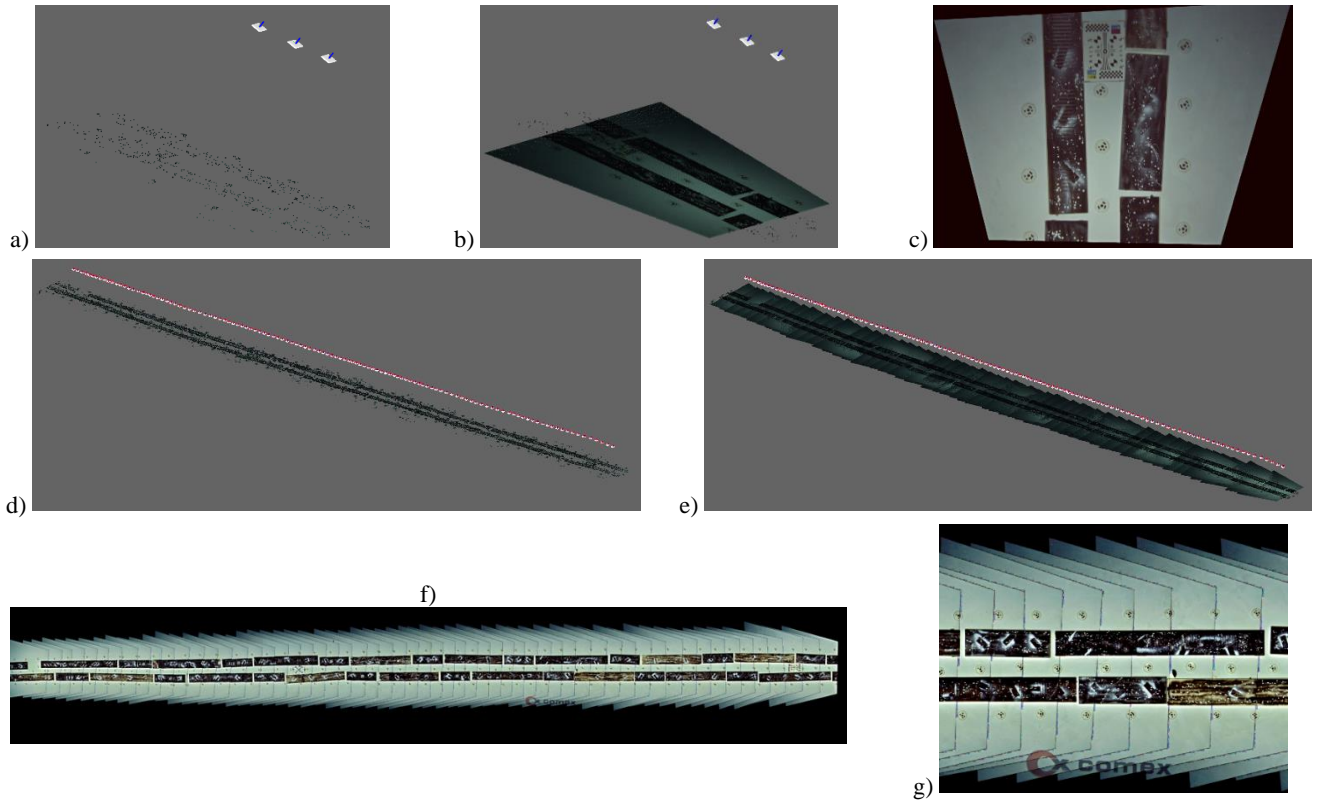

Figure 3. Results from the COMEX POOL dataset: first oriented frames with sparse point cloud (a), 3DSIP (b) and 2DSIM(c); entire image sequence oriented with sparse point cloud (c) and 3DSIP (d); final 2DSIM (f) with a detail (g).

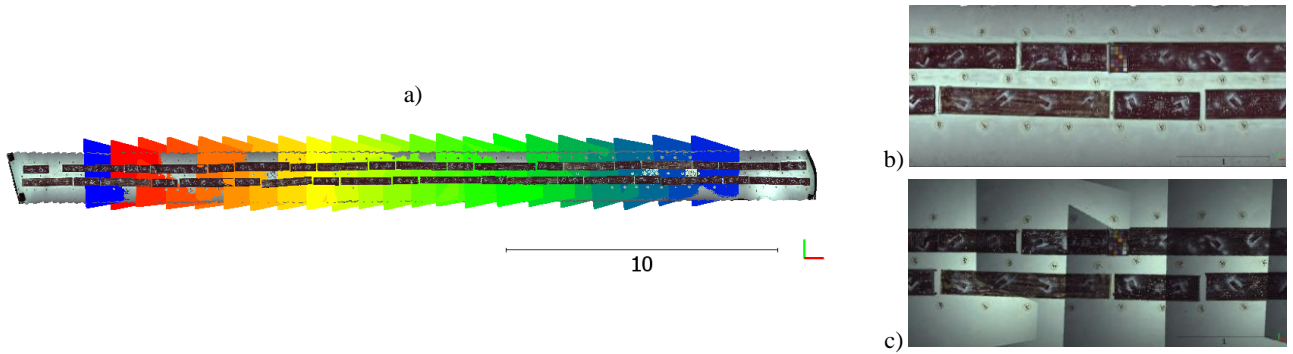

Figure 4. Accuracy assessment from the COMEX POOL dataset: 3DSIP point clouds over imposed in different colours to a 5-mm accurate textured mesh(a); detail of the textured mesh (b) and 3DSIP (c).

The procedure is tailored for future integration in other implementations, such as ORB-SLAM[1,2] and OpenVSLAM[3].

If a stereo or trifocal camera system is used (Drap et al., 2015b; Menna et al., 2019a and 2019b), data from one camera are selected for the following steps.

If AHRS data are collected and synchronised with the image recording, the attitude and heading readings are used to define the coordinated reference system of the survey.

For each image set defined by the wBA, the sparse point cloud is submitted to the plane fitting block and the pose of central image to mosaicing blocks 3.3 and 3.4.

### 3.3 The plane fitting module

The sparse point cloud of tie points is used to estimate the projection plane through a RANSAC best fitting approach implemented in Python. When the scene geometry is too far from the planarity assumption, the image plane projection step is skipped and only the sparse point cloud is visualised.

When AHRS information is available, the local horizontal plane passing through the centroid of the initialized map is chosen as XY plane; in the other cases the first plane fitted through the initialized map defines the XY plane for projection.

### 3.4 The sequential image stitching and image plane projection in object space module

The central image of each wBA subset is first undistorted (i.e. the lens distortions are removed according to the calibration parameters) and projected in the object space using the image pose and plane parameters computed at the previous steps. Two

---
[1] https://github.com/raulmur/ORB_SLAM
[2] https://github.com/raulmur/ORB_SLAM2
[3] https://github.com/xdspacelab/openvslam

different submodules are run. The two sub-blocks are currently implemented in Python and OpenCV.

**3.4.1  2DSIM - 2D sequential image mosaicing**: The images, warped according to the estimated homography (Figure 3c), are stitched together and the image mosaic is sequentially updated (Figure 3 f and g). The mosaic is re-initialized when the current local projection plane changes to a level (3D tie point residuals from the plane, or normal components) such that the validity assumptions for the stitching (i.e. planarity of the scene and/or purely rotational motion) are no longer verified. In this case a new plane is initialized, and a new image mosaic projection is started. At the same time, each single projected image is saved together with a separate ASCII world file containing the image-to-world (i.e., the local reference system defined in the previous steps) transformation. This allows the visualization of the georeferenced projected images in a GIS software.

**3.4.2  3DSIP - 3D sequential image plane projection**: A subsampled pixel number of the image plane (typically 150x100, as shown in 5.3) is projected in the object space in the form of a planar point cloud approximating the local surface (Figure 3b).

### 3.5  The 3D visualization module

Two different tools have been envisaged for rendering the 3D sparse point cloud together with camera positions and the 3D mosaics (3DSIP, Figure 3 d and e). For faster and more efficient visualization on a local scale, a WebSocket protocol is used to facilitate real-time data transfer to the remote computer (or tablet for a diver), which accesses the necessary 3D data for visualization trough a web browser. For large scale projects surveyed by ROVs using a support vessel, more powerful computational resources are typically available; in these cases, an incremental visualisation is carried out using the open-source WebGL based point cloud renderer Potree[4].

## 4. CASE STUDIES

The procedure detailed in section 3 is show-cased in several examples. The images were acquired with different platforms at different depths and in various underwater scenarios and seafloor characteristics, as summarised in table 1. Here, the real-time acquisition and processing is simulated as if the images were acquired by an ROV and transmitted to the surface computing unit, or as if the diver was equipped with an embedded device with separate processing units, one in charge of the image acquisition and computation of VO/SLAM/iSfM solution and the other of data visualization. In the presented experiments, the VO/SLAM/iSfM is run on images recorded at different the acquisition frame rate (Table 1). The incremental mosaicing, both in the image space (image mosaic) and object space (point cloud mosaic), is performed on reduced image resolutions. Experiments are realised considering both ¼ and 1/16 of the original images to assure a good compromise between costs and benefits in terms of both visualization and processing time.

### 4.1  Description of datasets

The case studies cover a broad range of application domains, environmental characteristics, and image acquisition platforms, i.e. underwater cameras operated by divers (COMEX POOL, NIWA T2-2017 and Moorea RESILIENCE SITE plots 4-7) and a submarine (XLENDI).

The COMEX POOL dataset, part of a series of subsea metrology qualification tests, was first presented in Menna et al. (2019a, 2019b) to investigate the use of vision-based real-time navigation and mapping techniques for underwater inspection and monitoring applications. The images were acquired with an ORUS3D 3Kv[5] system, in a controlled environment (the COMEX pool), at night to emulate typical operative high depths. The pool floor was covered with contrast plates to help the automatic image processing procedures.

The NIWA T2-2017 original video was recorded divers from the Antarctica New Zealand and NIWA (National Institute of Water and Atmospheric Research) in the Tethys Bay, Ross Sea, Antarctica. The dataset, as discussed in Piazza et al. (2018), was acquired within an ongoing international collaboration with a three-fold aim: (i) to extract 3D information from old video footage recorded following the common scientific diving practices, but not originally intended for photogrammetric purposes; (ii) to formulate best practices for 3D mapping and modelling of harsh underwater environments, such as in Antarctica; (iii) to develop automatic image sampling, processing and analysis procedures for non-destructive surveys of rocky-bottom benthic communities and habitats in Antarctica. The seafloor, mainly flat with some irregularities, is covered by living organisms, mainly sea stars, urchins.

The Moorea RESILIENCE SITE plots 4-7 dataset was acquired within the Moorea Island Digital Ecosystem Avatar (IDEA) project (https://mooreaidea.ethz.ch/), an inter-disciplinary and international project whose final aim is to digitize the entire Moorea island ecosystem (both inland and underwater) at different scales from island to microbes. Results of the photogrammetric underwater acquisitions and analyses were discussed in Neyer et al. (2018), Nocerino et al. (2019) and Rossi et al. (2019). The dataset was acquired by divers on a hilly bottom covered by coral reef.

The XLENDI dataset was acquired with the ROV-3D system (Drap et al., 2015b) mounted on a submarine (Drap et al., 2015a). It is a subset of a wider survey, realised within a deep underwater archaeological excavation over the Xlendi Phoenician shipwreck.

### 4.2  Results

The results of the different datasets are shown in Figures 3 to 7. It is here highlighted that the accuracy of navigation/orientation approach is not extensively investigated, since it is marginal to the focus of this study and mainly related to the choice of the VO/SLAM/iSfM.

As expected, the stitching approach in image space provides useful and satisfactory results in the case of a flat sea bottom such as in the case of the COMEX pool dataset (Figure 3 f and g). For accuracy assessment, the centre of circular targets fixed on the pool from the 3DSIP point clouds are compared with reference 3D coordinates measured with a laser tracker (see Menna et al., 2019a and 2019b for details). The RMS of the similarity transformation residuals results below five centimetres. In Figure 4a 3DSIP point clouds are over imposed in different colours to a 5-mm accurate textured mesh. Details of the textured mesh and 3DSIP are shown for visual comparison in Figure 4b and 4c, respectively.

In all the other investigated scenarios, where the seafloor does not fulfil the planarity condition, the image mosaicing is automatically split in smaller parts (2DSIM in Figures 5d, 5e, 6e, 6f, 7d, 7e).

---

[4] http://potree.org/

[5] https://comex.fr/en/orus3d/

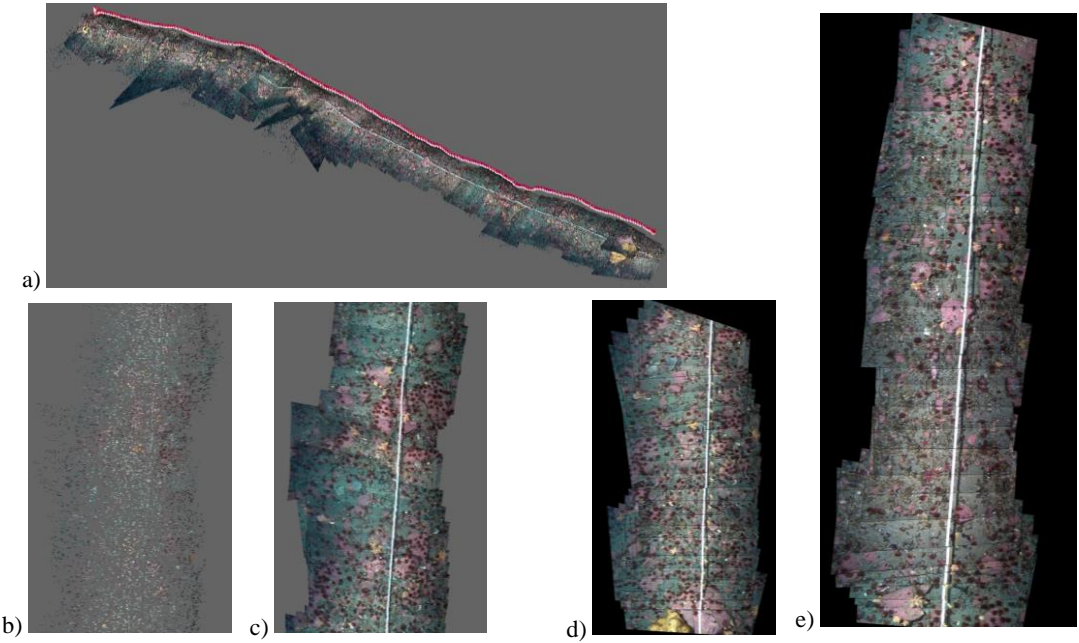

Figure 5. Results from the NIWA T2-2017 dataset: entire image sequence oriented with sparse point cloud and 3DSIP (a); details of the sparse point cloud (b), and 3DSIP (c); two separate parts of the 2DSIM (d, e).

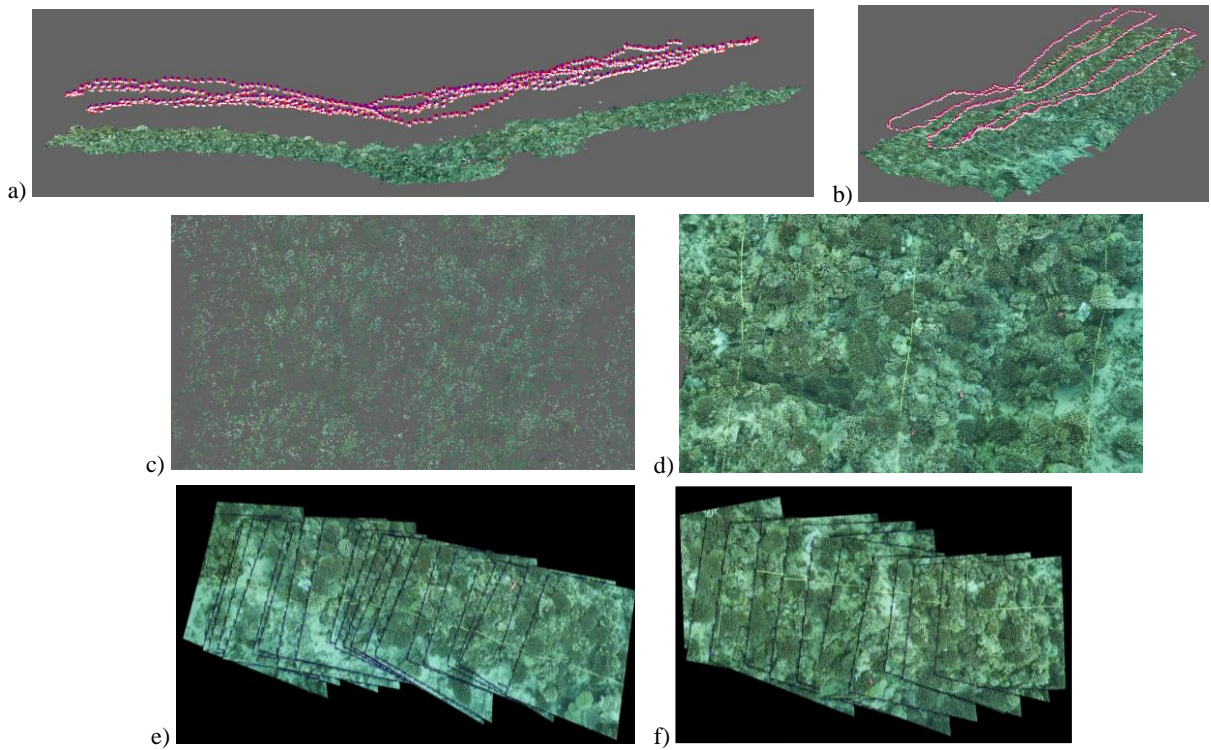

Figure 6. Results from the Moorea RESILIENCE SITE plots 4-7 dataset: first oriented frames with sparse point cloud (a) and 3DSIP (b); details of the sparse point cloud (c), and 3DSIP (d); two separate parts of the 2DSIP (e, f).

On the contrary, the incremental 3DSIP follows the seabed topography (Figures 5a, 6a, 6b, 7a), providing a much denser and clearer idea of the area incrementally covered during the survey (Figures 4b, 4d, 5c, 6d, 7c) with respect to the visualization of the sparse point cloud (Figures 3a, 3c, 5b, 6c, 7b).

## 5. DISCUSSION AND OUTLOOK

In this paper, a visual aid for real-time underwater navigation and mapping has been presented. Primarily tailored to support the piloting of ROVs, the method can be extended and integrated into an embedded image-based surveying system employed by divers. The method is in fact modular and implemented in four building blocks, each of them designed to run on different computing units. The first block integrates the image-based navigation and mapping module, whose outputs are employed to estimate the reference plane to build an incremental image mosaic, both in 2D and in 3D, in the subsequent module.

While the 2D mosaic, or 2DSIM, is kept in background, the user can visualize in real-time the updating 3D scene composed of the camera positions, sparse point cloud and 3D mosaic or 3DSIP. The accuracy of visualization is a function of the object roughness and thus local variations from least square plane estimated with RANSAC procedure. Therefore, for objects where the local plane approximation is valid, the accuracy is solely dependent on the VO/SLAM method utilised. The procedure has been successfully tested with data collected in different real-case scenarios. The developed procedure can be integrated in a modular system architecture for underwater photogrammetry where the image mosaicing procedures 2DSIM and 3DSIP are distributed on different computers. For example, 3DSIP can be performed underwater at 0.5 Hertz even using low cost computers (e.g. such as Raspberry Pi4) that is acceptable for real time navigation on a compact submersed system controlled via an underwater tablet by divers. On more demanding applications, such as in the case of subsea metrology inspections, powerful computers are typically available in a distributed architecture (for example the COMEX ORUS3D, https://comex.fr/en/orus3d/) and the proposed technique can integrate both 3DSIP and 2DSIM at higher frequency with virtually no limits with respect to the maximum number of points rendered.

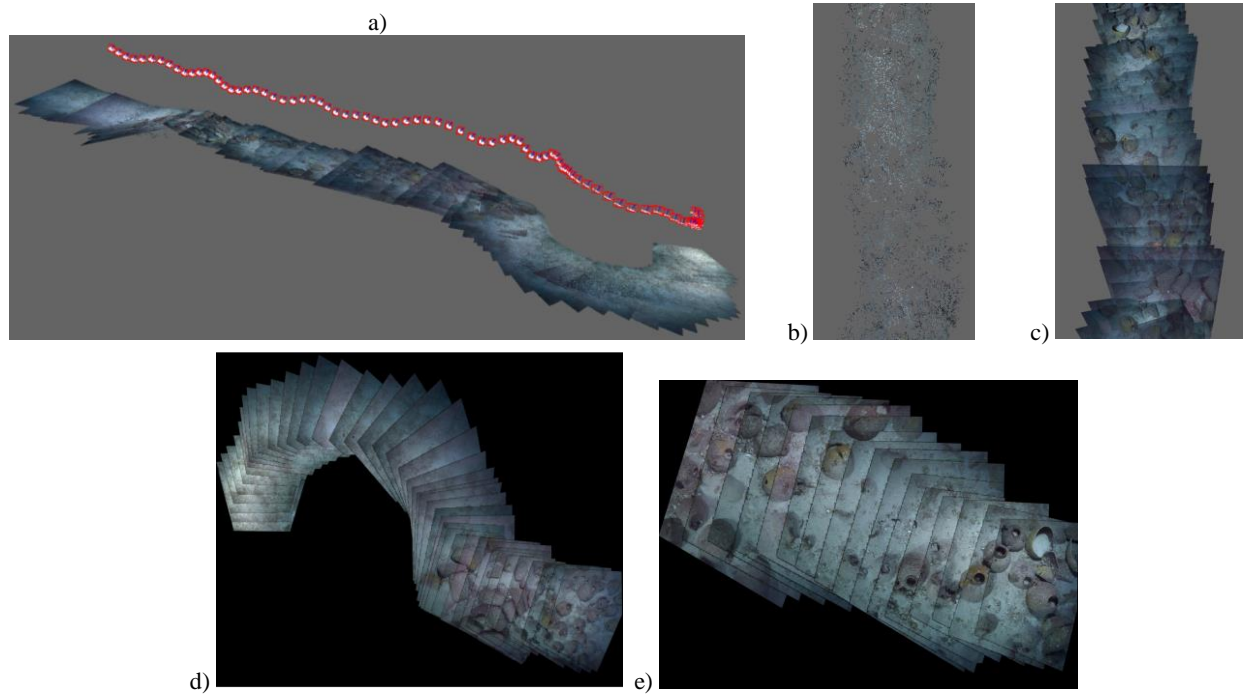

Figure 7. Results from the XLENDI dataset: entire image sequence oriented with sparse point cloud and 3DSIP (a); details of the sparse point cloud (b), and 3DSIP (c); two separate parts of the 2DSIP (d, e).

## ACKNOWLEDGEMENTS

This study was partially supported by the French Single Inter-Ministry Fund (FUI), in the framework of the GREENEXPLORER project.
The COMEX POOL dataset was acquired by COMEX (COmpanie Maritime d'EXpertise Marseille, France) within the GREENEXPLORER project.
The NIWA T2-2017 dataset acquisition was funded by NZ Ministry of Primary Industries and Ministry of Business, Innovation and Employment, respectively, and supported by Antarctica New Zealand and the excellent NIWA dive teams. It was processed within the Project "ICE-LAPSE" (PNRA 2013/AZ1.16: "Analysis of Antarctic benthos dynamics by using non-destructive monitoring devices and permanent stations"), funded by the Italian National Antarctic Program. The video recording was performed by Ian Hawes (University of Waikato, Waikato, NZ).
The Moorea RESILIENCE SITE plots 4-7 dataset was acquired within the Moorea Island Digital Ecosystem Avatar (IDEA) project, supported by the financial and scientific support of Prof. Matthias Troyer through the Institute of Theoretical Physics, ETH Zurich, the U.S. National Science Foundation under Grant No. OCE 16-37396 (and earlier awards) as well as a generous gift from the Gordon and Betty Moore Foundation. The IDEA project is executed under permits issued by the French Polynesian Government (Délégation à la Recherche) and the Haut-Commissariat de la République en Polynésie Francaise (DTRT) (Protocole d'Accueil 2005-2018). The authors are grateful to Dr. A. Brooks (Marine Science Institute, University of California, Santa Barbara - UCSB, California, USA), Prof. A. Capra (DIEF Department, University of Modena and Reggio Emilia, Modena, Italy) and the UCSB Gump station team in Moorea for their crucial support in the data acquisition, and Prof. A. Gruen (Institute of Theoretical Physics, ETH Zurich) for his scientific support.
The XLENDI dataset was acquired by COMEX the framework of the GROPLAN project (http://www.lsis.org/groplan/), funded by the National Agency of Research (ANR) of France. The XLENDI project is carried out with the support of the Superintendence of Cultural Heritage of Malta and lead by Prof. T. Gambin (Department of Classics and Archaeology, University of Malta) and Prof. J.-C. Sourisseau (Aix-Marseille University, Centre Camille Jullian, CNRS, UMR 7299, Aix-En-Provence, France).
A special thank goes to Alessandro Torresani and Daniele Morabito from 3DOM FBK for the useful insights on the real-time visualization tool architecture.